\documentclass[runningheads]{llncs}
\usepackage{graphicx}
\usepackage{amsmath,amssymb} 
\usepackage{color}
\usepackage[width=122mm,left=12mm,paperwidth=146mm,height=193mm,top=12mm,paperheight=217mm]{geometry}
\usepackage{caption}
\usepackage{url}

\usepackage[normalem]{ulem}
\usepackage[table,xcdraw]{xcolor}
\usepackage{array}
\usepackage{tabularx}

\usepackage{fancybox}
\usepackage{amsfonts,amsmath,amssymb,latexsym}
\usepackage{array}
\usepackage{graphicx}

\newenvironment{separate}{\begin{quote}\begin{sffamily}\footnotesize}{\end{sffamily}\end{quote}}

\newcommand{\beq}{\begin{equation}}
\newcommand{\eeq}{\end{equation}}
\newcommand{\beqa}{\begin{eqnarray}}
\newcommand{\eeqa}{\end{eqnarray}}
\newcommand{\beqas}{\begin{eqnarray*}}
\newcommand{\eeqas}{\end{eqnarray*}}

\newlength{\boxwidth}
\newlength{\narrowboxwidth}
\usepackage{upquote}

\newif\ifcomments
\commentstrue

\newif\iffinal
\finaltrue

\iffinal
\commentstrue
\else\fi

\ifcomments
\newcommand{\ct}[1]{\textcolor{magenta}{\textsc{CT}: #1}}
\newcommand{\rc}[1]{\textcolor{cyan}{\textsc{RC}: #1}}
\newcommand{\fs}[1]{\textcolor{red}{\textsc{FS}: #1}}
\newcommand{\er}[1]{\textcolor{blue}{\textsc{ER}: #1}}
\newcommand{\rz}[1]{\textcolor{green}{\textsc{RZ}: #1}}
\else
\newcommand{\ct}[1]{}
\newcommand{\rc}[1]{}
\newcommand{\fs}[1]{}
\newcommand{\er}[1]{}
\newcommand{\rz}[1]{}
\fi

\begin{document}
\pagestyle{headings}
\mainmatter
\def\ECCV16SubNumber{3}  

\title{Performance Measures and a Data Set for Multi-Target, Multi-Camera Tracking} 
\iffinal
	\renewcommand{\thefootnote}{$\star$}
	\footnotetext{This material is based upon work supported by the National Science Foundation
		under grants CCF-1513816 and IIS-1543720 and by the Army Research Office under grant W911NF-16-1-0392.}
\else
\fi

\titlerunning{Performance Measures and a Data Set for MTMC Tracking}

\authorrunning{Ergys Ristani, Francesco Solera, Roger S. Zou, Rita Cucchiara, Carlo Tomasi}

\iffinal
	\author{Ergys Ristani$^{1}$, Francesco Solera$^{2}$, Roger S. Zou$^{1}$,\\Rita Cucchiara$^{2}$, and Carlo Tomasi$^{1}$}
	\institute{$^{1}$ Computer Science Department, Duke University, Durham, USA\\
$^{2}$ Department of Engineering, University of Modena and Reggio Emilia, Modena, Italy} 
\else
	\author{Anonymous BMTT submission}
	\institute{Paper ID \ECCV16SubNumber}
\fi

\maketitle

\begin{abstract}
To help accelerate progress in multi-target, multi-camera tracking systems, we present (i) a new pair of precision-recall measures of performance that treats errors of all types uniformly and emphasizes correct identification over sources of error; (ii) the largest fully-annotated and calibrated data set to date with more than 2 million frames of 1080p, 60fps video taken by 8 cameras observing more than 2,700 identities over 85 minutes; and (iii) a reference software system as a comparison baseline. We show that (i) our measures properly account for bottom-line identity match performance in the multi-camera setting; (ii) our data set poses realistic challenges to current trackers; and (iii) the performance of our system is comparable to the state of the art. 
\end{abstract}

\section{Introduction}
\emph{Multi-Target, Multi-Camera} (MTMC) tracking systems automatically track multiple people through a network of cameras.
As MTMC methods solve larger and larger problems, it becomes increasingly important (i) to agree on straightforward performance measures that consistently report bottom-line tracker performance, both within and across cameras; (ii) to develop realistically large benchmark data sets for performance evaluation; and (iii) to compare system performance end-to-end. This paper contributes to these aspects.

\noindent\textbf{Performance Measures.}
Multi-Target Tracking has been traditionally defined as continuously following multiple objects of interest. Because of this, existing performance measures such as CLEAR MOT report how often a tracker makes what types of incorrect decisions. We argue that some system users may instead be more interested in how well they can determine who is where at all times.

To see this distinction, consider the scenario abstractly depicted in Figure~\ref{fig:frag}(a) and~\ref{fig:frag}(c). Airport security is following suspect A spotted in the airport lobby. They need to choose between two trackers, 1(a) and 1(c). Both tag the suspect as identity 1 and track him up to the security checkpoint. System 1(a) makes a single mistake at the checkpoint and henceforth tags the suspect as identity 2, so it loses the suspect at the checkpoint. After the checkpoint, system 1(c) repeatedly flips the tags for suspect A between 1 and 2, thereby giving police the correct location of the suspect several times also between the checkpoint and the gate, and for a greater overall fraction of the time. Even though system 1(a) incurs only one ID switch, airport security is likely to prefer system 1(c), which reports the suspect's position longer---multiple ID switches notwithstanding---and ultimately leads to his arrest at the gate.

We do not claim that one measure is better than the other, but rather that different measures serve different purposes. \emph{Event-based} measures like CLEAR MOT help pinpoint the source of some errors, and are thereby informative for the designer of certain system components. In the interest of users in applications such as sports, security, or surveillance, where preserving identity is crucial, we propose two \emph{identity-based} measures (ID precision and ID recall) that evaluate how well computed identities conform to true identities, while disregarding where or why mistakes occur. Our measures apply both within and across cameras.





\noindent\textbf{Data Set.}
We make available a new data set that has more than 2 million frames and more than 2,700 identities. It consists of $8\times 85$ minutes of 1080p video recorded at 60 frames per second from 8 static cameras deployed on the Duke University campus during periods between lectures, when pedestrian traffic is heavy. Calibration data determines homographies between images and the world ground plane. All trajectories were manually annotated by five people over a year, using an interface we developed to mark trajectory key points and associate identities across cameras. The resulting nearly 100,000 key points were automatically interpolated to single frames, so that every identity comes with single-frame bounding boxes and ground-plane world coordinates across all cameras in which it appears. To our knowledge this is the first dataset of its kind. 

\noindent\textbf{Reference System.}
We provide code for an MTMC tracker that extends a single-camera system that has shown good performance~\cite{ristani2014tracking} to the multi-camera setting. We hope that the conceptual simplicity of our system will encourage plug-and-play experimentation when new individual components are proposed.

We show that our system does well on a recently published data set~\cite{CaoCCZH15} when previously used measures are employed to compare our system to the state of the art. This comparison is only circumstantial because most existing results on MTMC tracking report performance using \emph{ground-truth} person detections and \emph{ground-truth} single-camera trajectories as inputs, rather than using the results from \emph{actual} detectors and single-camera trackers. The literature typically justifies this limitation with the desire to measure only what a multi-camera tracker \emph{adds} to a single-camera system. This justification is starting to wane as MTMC tracking systems approach realistically useful performance levels. Accordingly, we evaluate our system end-to-end, and also provide our own measures as a baseline for future research.


\section{Related Work}
\label{sec:literature}

We survey prior work on MTMC performance measures, data sets, and trackers.

\noindent\textbf{Measures.} We rephrase existing MTMC performance measures as follows.

\begin{itemize}
    \item A \emph{fragmentation} occurs in frame $t$ if the tracker switches the identity of a trajectory in that frame, but the corresponding ground-truth identity does not change. The number of fragmentations at frame $t$ is $\phi_t$, and $\Phi = \sum_t \phi_t$.
    \item A \emph{merge} is the reverse of a fragmentation: The tracker merges two different ground truth identities into one between frames $t'$ and $t$. The number of merges at frame $t$ is $\gamma_t$, and $\Gamma = \sum_t \gamma_t$.
    \item A \emph{mismatch} is either a fragmentation or a merge. We define $\mu_t = \phi_t + \gamma_t$ and $M = \sum_t \mu_t$. 
\end{itemize}
When relevant, each of these error counts is given a superscript $w$ (for ``within-camera'') when the frames $t'$ and $t$ in question come from the same camera, and a superscript $h$ (for ``handover'') otherwise.

The number of \emph{false positives} $fp_t$ is the number of times the tracker detects a target in frame $t$ where there is none in the ground truth, the number of \emph{false negatives} $fn_t$ is the number of true targets missed by the tracker in frame $t$, and $tp_t$ is the number of true positive detections at time $t$. The capitalized versions $TP$, $FP$, $FN$ are the sums of $tp_t$, $fp_t$, and $fn_t$ over all frames (and cameras, if more than one), and the superscripts $w$ and $h$ apply here as well if needed.

\emph{Precision} and \emph{recall} are the usual derived measures, $P = TP/(TP + FP)$ and $R = TP/(TP + FN)$.

Single-camera, multi-object tracking performance is typically measured by the Multiple Object Tracking Accuracy (MOTA):
\begin{equation}
\label{eq:mota}
\text{MOTA} = 1-\frac{FN + FP + \Phi}{T}
\end{equation}
and related scores (MOTP, MT, ML, FRG)~\cite{bernardin08,wu2006tracking,milan2013challenges}. MOTA penalizes detection errors ($FN+FP$) and fragmentations ($\Phi$) normalized by the total number $T$ of true detections. If extended to the multi-camera case, MOTA and its companions under-report across-camera errors, because a trajectory that covers $n_f$ frames from $n_c$ cameras has only about $n_c$ across-camera detection links between consecutive frames and about $n_f - n_c$ within camera ones, and $n_c \ll n_f$. To address this limitation handover errors~\cite{kuo_intercamera_2010} and multi-camera object tracking accuracy (MCTA)~\cite{CaoCCZH15,mctchallenge} measures were introduced, which we describe next.

\noindent\textit{Handover errors} focus only on errors across cameras, and distinguish between fragmentations $\Phi^h$ and merges $\Gamma^h$. Fragmentations and merges are divided further into crossing ($\Phi^h_X$ and $\Gamma^h_X$) and returning ($\Phi^h_R$ and $\Gamma^h_R$) errors.  These more detailed handover error scores help understand different types of tracker failures, and within-camera errors are quantified separately by standard measures.

\noindent\textit{MCTA} condenses all aspects of system performance into one measure:
\begin{equation}
\text{MCTA} = \underbrace{\vphantom{\left(\frac{1}{1}\right)}\frac{2PR}{P+R}}_{F_1}\underbrace{\left(1-\frac{M^w}{T^w}\right)}_{\text{within camera}}\underbrace{\left(1-\frac{M^h}{T^h}\right)}_{\text{handover}} \;.
\label{eq:MCTA}
\end{equation}
This measure multiplies the $F_1$ detection score (harmonic mean of precision and recall) by a term that penalizes within-camera identity mismatches ($M^w$) normalized by true within-camera detections ($T^w$) and a term that penalizes wrong identity handover mismatches ($M^h$) normalized by the total number of handovers. Consistent with our notation, $T^h$ is the number of true detections (true positives $TP^h$ plus false negatives $FN^h$) that occur when consecutive frames come from different cameras.

Comparing to MOTA, MCTA multiplies within-camera and handover mismatches rather than adding them. In addition, false positives and false negatives, accounted for in precision and recall, are also factored into MCTA through a product. This separation brings the measure into the range $[0, 1]$ rather than $[-\infty, 1]$ as for MOTA. However, the reasons for using a product rather than some other form of combination are unclear. In particular, each error in any of the three terms is penalized inconsistently, in that its cost is multiplied by the (variable) product of the other two terms.

\noindent\textbf{Data Sets.}
Existing multi-camera data sets allow only for limited evaluation of MTMC systems. Some have fully overlapping views and are restricted to short time intervals and controlled conditions~\cite{Fleuret08a,BerclazFTF11,ferryman2009overview}. Some sports scenarios provide quality video with many cameras~\cite{d2009semi,de2008distributed}, but their environments are severely constrained and there are no blind spots between cameras. Data sets with disjoint views come either with low resolution video~\cite{CaoCCZH15,kuo_intercamera_2010,zhang_camera_2015}, a small number of cameras placed along a straight path~\cite{CaoCCZH15,kuo_intercamera_2010}, or scripted scenarios~\cite{CaoCCZH15,Fleuret08a,BerclazFTF11,ferryman2009overview,zhang_camera_2015,per2012dana36}. Most importantly, all existing data sets only have a small number of identities. Table \ref{tab:data sets} summarizes the parameters of existing data sets. Ours is shown in the last row. It contains more identities than all previous data sets \emph{combined}, and was recorded over the longest time period at the highest temporal resolution (60 fps).

\begin{table}[t]
\centering
{\tiny
\begin{tabular}{|l|c|c|c|c|c|c|c|c|c|c|c|}
\hline
\bf Dataset                                         & \bf IDs   & \bf Duration & \bf Cams & \bf Actors &\bf  Overlap &\bf  Blind Spots  & \bf Calib. & \bf Resolution &\bf  FPS &\bf  Scene &\bf  Year \\ \hline
Laboratory~\cite{Fleuret08a}                    & 3     & 2.5 min  & 4    & Yes    & Yes     & No          & Yes         & 320x240    & 25  & Indoor      & 2008 \\
Campus~\cite{Fleuret08a}                        & 4     & 5.5 min  & 3    & Yes    & Yes     & No          & Yes         & 320x240    & 25  & Outdoor     & 2008 \\
Terrace~\cite{Fleuret08a}                       & 7     & 3.5 min  & 4    & Yes    & Yes     & No          & Yes         & 320x240    & 25  & Outdoor     & 2008 \\
Passageway~\cite{BerclazFTF11}                  & 4     & 20 min   & 4    & Yes    & Yes     & No          & Yes         & 320x240    & 25  & Mixed       & 2011 \\
Issia Soccer~\cite{d2009semi}                   & 25    & 2 min    & 6    & No     & Yes     & No          & Yes         & 1920x1080  & 25  & Outdoor     & 2009 \\  
Apidis Basket.~\cite{de2008distributed}      & 12    & 1 min    & 7    & No     & Yes     & No          & Yes         & 1600x1200  & 22  & Indoor      & 2008 \\
PETS2009~\cite{ferryman2009overview}            & 30    & 1 min    & 8    & Yes    & Yes     & No          & Yes         & 768x576    & 7   & Outdoor      & 2009 \\
NLPR MCT 1~\cite{CaoCCZH15}                     & 235   & 20 min   & 3    & No     & No      & Yes         & No          & 320x240    & 20  & Mixed       & 2015 \\
NLPR MCT 2~\cite{CaoCCZH15}                     & 255   & 20 min   & 3    & No     & No      & Yes         & No          & 320x240    & 20  & Mixed       & 2015 \\
NLPR MCT 3~\cite{CaoCCZH15}                     & 14    & 4 min    & 4    & Yes    & Yes     & Yes         & No          & 320x240    & 25  & Indoor      & 2015 \\
NLPR MCT 4~\cite{CaoCCZH15}                     & 49    & 25min    & 5    & Yes    & Yes     & Yes         & No          & 320x240    & 25  & Mixed       & 2015 \\
Dana36~\cite{per2012dana36}                     & 24    & N/A      & 36   & Yes    & Yes     & Yes         & No          & 2048x1536  & N/A & Mixed       & 2012 \\ 
USC Campus~\cite{kuo_intercamera_2010}          & 146   & 25 min   & 3    & No     & No      & Yes         & No          & 852x480    & 30  & Outdoor     & 2010 \\
CamNeT~\cite{zhang_camera_2015}                 & 50    & 30 min   & 8    & Yes    & Yes     & Yes         & No          & 640x480    & 25  & Mixed       & 2015 \\
DukeMTMC (ours)                                 & 2834  & 85 min   & 8    & No     & Yes     & Yes         & Yes         & 1920x1080  & 60  & Outdoor     & 2016 \\
\hline
\end{tabular}}
\caption{Summary of existing data sets for MTMC tracking. Ours is in the last row.}
\label{tab:data sets}
\vspace{-1cm}
\end{table}

\noindent\textbf{Systems.}
MTMC trackers rely on pedestrian detection~\cite{Benenson_2014} and tracking~\cite{MOTChallenge2015} or assume single-camera trajectories to be given~\cite{kuo_intercamera_2010,zhang_camera_2015,bredereck_data_2012,cai_exploring_2014,chen_adaptive_2011,chen_multitarget_2015,chen_direction-based_2011,daliyot_framework_2013,das2014consistent,gilbert_tracking_2006,javed_modeling_2008,makris_bridging_2004}. 
\textit{Spatial relations between cameras} are either explicitly mapped in 3D~\cite{zhang_camera_2015,chen_adaptive_2011}, learned by tracking known identities~\cite{javed_modeling_2008,jiuqing_distributed_2013,4407431}, or obtained by comparing entry/exit rates across pairs of cameras~\cite{kuo_intercamera_2010,cai_exploring_2014,makris_bridging_2004}.
Pre-processing methods may fuse data from partially overlapping views~\cite{zhang_tracking_2015}, while some systems rely on completely overlapping and unobstructed views~\cite{BerclazFTF11,bredereck_data_2012,ayazoglu_dynamic_2011,kamal_information_2013,hamid_player_2010}. People \textit{entry and exit points} may be explicitly modeled on the ground~\cite{kuo_intercamera_2010,cai_exploring_2014,chen_adaptive_2011,makris_bridging_2004} or image plane~\cite{gilbert_tracking_2006,jiuqing_distributed_2013}. 
\textit{Travel time} is also modeled, either parametrically~\cite{zhang_camera_2015,jiuqing_distributed_2013} or not~\cite{kuo_intercamera_2010,chen_adaptive_2011,gilbert_tracking_2006,javed_modeling_2008,makris_bridging_2004}.

\noindent \textit{Appearance} is captured by color ~\cite{kuo_intercamera_2010,zhang_camera_2015,cai_exploring_2014,chen_adaptive_2011,chen_multitarget_2015,chen_direction-based_2011,das2014consistent,gilbert_tracking_2006,javed_modeling_2008,zhang_tracking_2015,jiuqing_distributed_2013} 
and texture descriptors~\cite{kuo_intercamera_2010,zhang_camera_2015,cai_exploring_2014,chen_multitarget_2015,daliyot_framework_2013,zhang_tracking_2015}. Lighting variations are addressed through color normalization~\cite{cai_exploring_2014}, exemplar based approaches~\cite{chen_multitarget_2015}, or brightness transfer functions learned with~\cite{das2014consistent,javed_modeling_2008} or without supervision~\cite{zhang_camera_2015,chen_adaptive_2011,gilbert_tracking_2006,zhang_tracking_2015}.
Discriminative power is improved by \emph{saliency} information~\cite{martinel2014saliency,zhao2013unsupervised} or \emph{learning} features specific to body parts~\cite{kuo_intercamera_2010,cai_exploring_2014,chen_multitarget_2015,chen_direction-based_2011,daliyot_framework_2013,das2014consistent,jiuqing_distributed_2013}, either in the image~\cite{BedagkarGala6130457,BedagkarGala20121908,ChengBMVC2568} or back-projected onto an articulated~\cite{learning_baltieri_2013,Shaogang14} or monolithic~\cite{mapping_baltieri_2015} 3D body model.

\noindent All MTMC trackers employ \emph{optimization} to maximize the coherence of observations for predicted identities. They first summarize spatial, temporal, and appearance information into a graph of \emph{weights} $w_{ij}$ that express the affinity of node observations $i$ and $j$, and then partition the nodes into identities either greedily through bipartite matching or, more generally, by finding either paths or cliques with maximal internal weights. Some contributions are as follows:
\begin{table}
\vspace{-0.5cm}
\centerline{\setlength{\tabcolsep}{1.2mm}\noindent\begin{tabular}{l|p{3.0cm}|p{2.2cm}|p{1.1cm}}
 & \multicolumn{1}{c|}{Single-Camera} & \multicolumn{1}{c|}{Cross-Camera} & \multicolumn{1}{c}{Both} \\\hline
Bipartite & \cite{brendel2011multiobject,shu2012part,wu2007detection} &
\cite{kuo_intercamera_2010,cai_exploring_2014,chen_multitarget_2015,daliyot_framework_2013} & --- \\\hline
Path & \cite{BerclazFTF11,Izadinia_ECCV12_MP2T,pirsiavash2011globally,zhang2008global} & \cite{javed_modeling_2008,jiuqing_distributed_2013} & \cite{CaoCCZH15,zhang_tracking_2015}\\\hline
Clique &
\cite{ristani2014tracking,butt2013multiple,chari2015pairwise,collins2012multitarget,dehghan2015gmmcp,kumar2014multiple,shafique2005noniterative,tang2015subgraph,wen2014multiple,ZamirECCV12} &
\cite{das2014consistent} & Ours
\end{tabular}}
\caption{Optimization techniques employed by MTMC systems.}
\end{table}
\vspace{-1cm}

\noindent In this paper, we extend a previous clique method~\cite{ristani2014tracking} to formulate within- and across-camera tracking in a unified framework, similarly to previous MTMC flow methods~\cite{CaoCCZH15,zhang_tracking_2015}. In contrast with \cite{das2014consistent}, we handle identities reappearing in the same camera and differently from \cite{Fleuret08a,BerclazFTF11} we handle co-occuring observations in overlapping views naturally, with no need for separate data fusion methods. 


\section{Performance Measures}
\label{sec:performance_measures}

Current event-based MTMC tracking performance measures count mismatches between ground truth and system output through \emph{changes} of identity over time. The next two Sections show that this can be problematic both within and across cameras. The Section thereafter introduces our proposed measures.

\subsection{Within-Camera Issues}

\begin{figure}[t]
    \centering
    \begin{tabular}{ccc}
    \includegraphics[width=0.3\textwidth]{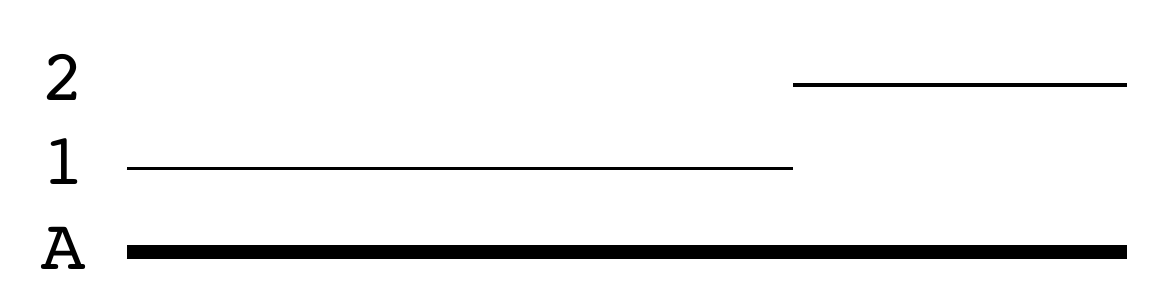} &
    \includegraphics[width=0.3\textwidth]{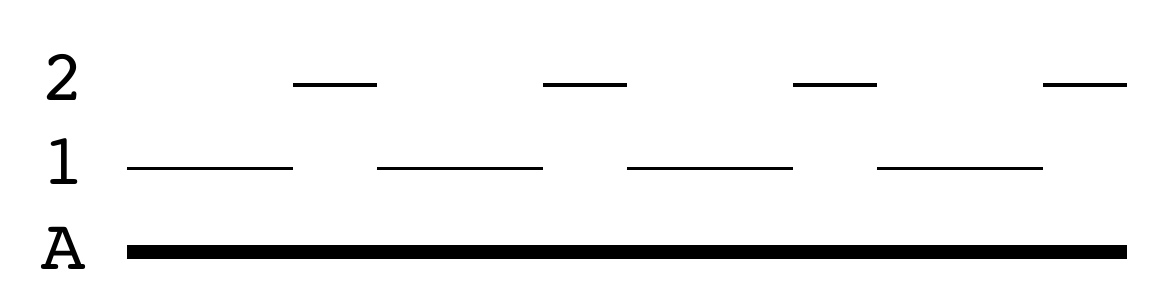} &
    \includegraphics[width=0.3\textwidth]{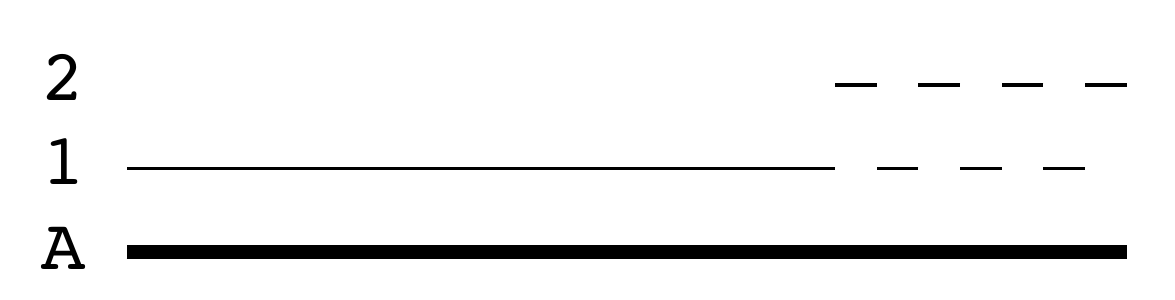}\\
    (a) & (b) & (c)
    \end{tabular}
    \caption{Where there is one true identity \texttt{A} (thick line, with time in the horizontal direction), a tracker may mistakenly compute identities \texttt{1} and \texttt{2} (thin lines) broken into two fragments (a) or into eight (b, c). Identity \texttt{1} covers 67\% of the true identity's trajectory in (a) and (b), and 83\% of it in (c). Current measures charge one fragmentation error to (a) and 7 to each of (b) and (c). Our proposed measure charges 33\% of the length of \texttt{A} to each of (a) and (b), and 17\% to (c).}
    \label{fig:frag}
\end{figure}

With event-based measures, a truly-unique trajectory that switches between two computed identities over $n$ frames can incur penalties that are anywhere between 1, when there is exactly one switch, and $n-1$, in the extreme case of one identity switch per frame. This can yield inconsistencies if correct identities are crucial. For example, in all cases in Figure \ref{fig:frag}, the tracker covers a true identity \texttt{A} with computed identities \texttt{1} and \texttt{2}. Current measures would make cases (b) and (c) equally bad, and (a) much better than the other two.

And yet the key mistake made by the tracker is to see two identities where there is one. To quantify the extent of the mistake, we need to decide which of the two computed identities we should match with \texttt{A} for the purpose of performance evaluation. Once that choice is made, every frame in which \texttt{A} is assigned to the wrong computed identity is a frame in which the tracker is in error.

Since the evaluator---and not the tracker---makes this choice, we suggest that it should favor the tracker to the extent possible. If this is done for each tracker under evaluation, the choice is fair. In all cases in Figure \ref{fig:frag}, the most favorable choice is to tie \texttt{A} to \texttt{1}, because this choice explains the largest fraction of \texttt{A}.

Once this choice is made, we measure the number of frames over which the tracker is wrong---in the example, the number of frames of \texttt{A} that are not matched to \texttt{1}. In Figure \ref{fig:frag}, this measure makes (a) and (b) equally good, and (c) better than either. This penalty is consistent because it reflects precisely what the choice made above maximizes, namely, the number of frames over which the tracker is correct about who is where. In (a) and (b), the tracker matches ground truth 67\% of the time, and in (c) it matches it 83\% of the time.

Figure \ref{fig:frag} is about fragmentation errors. It can be reinterpreted in terms of merge errors by exchanging the role of thick and thin lines. In this new interpretation, choosing the longest ground-truth trajectory as the correct match for a given computed trajectory explains as much of the tracker's output as possible, rather than as much of the ground truth. In both directions, our \emph{truth-to-result matching} criterion is to let ground truth and tracker output explain as much of each other's data as possible, in a way that will be made quantitative later on.

\subsection{Handover Issues}

\begin{figure}
    \centering
    \begin{tabular}{ccc}
    \includegraphics[width=0.45\textwidth]{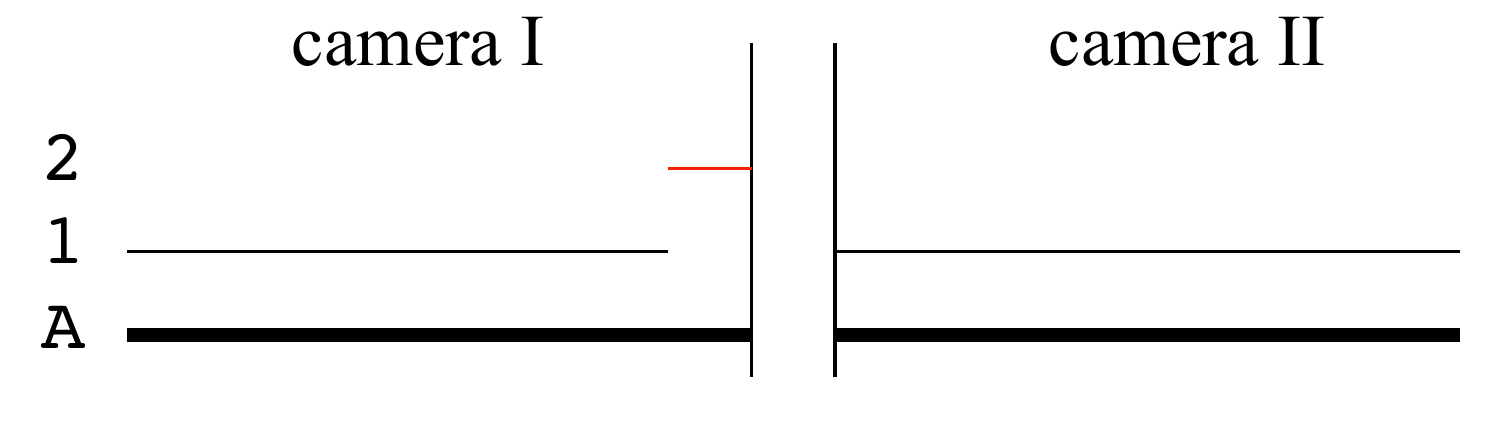} & \phantom{wide} &
    \includegraphics[width=0.45\textwidth]{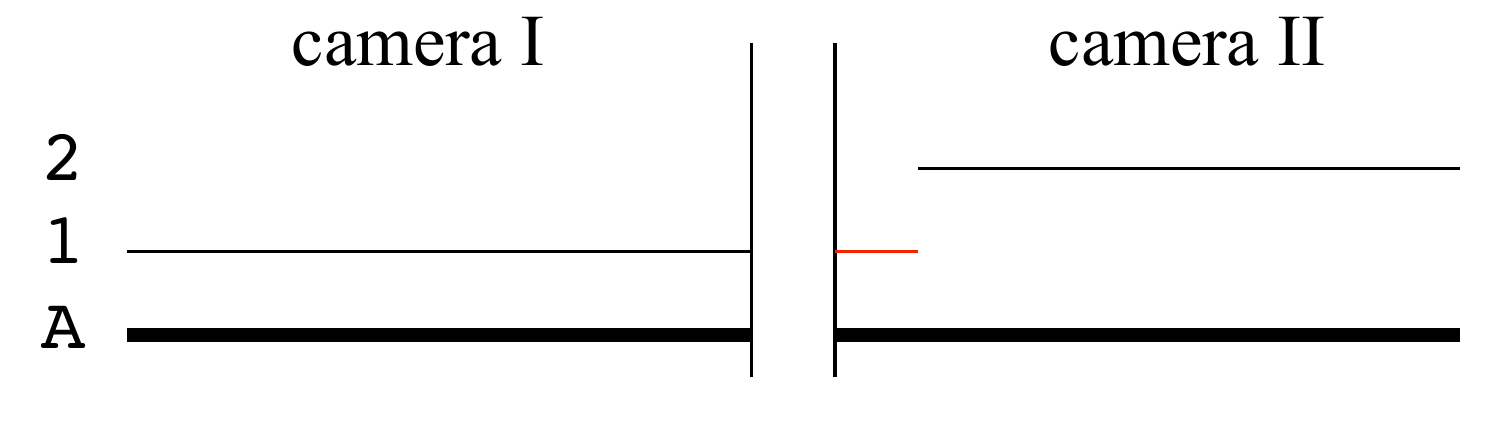}\\
    (a) & & (b)
    \end{tabular}
    \caption{(a) Ground-truth trajectory \texttt{A} is handed over correctly between cameras, because it is given the same computed identity \texttt{1} throughout, except that  a short fragment in camera I is mistakenly given identity \texttt{2} (red). This counts as a handover error with existing measures. (b) \texttt{A} is handed over incorrectly, but a short fragment in camera II mistakenly given identity \texttt{1} (red) makes existing measures \emph{not} count it as a handover error. Existing measures would charge a within-camera fragmentation and an across-camera fragmentation to (a) and one within-camera fragmentation to (b), even if assignment (a) is much better than (b) in terms of target identification.}
    \label{fig:handover}
\end{figure}

Event-based measures often evaluate handover errors separately from within-camera errors: Whether a mismatch is within-camera or handover depends on the identities associated to the very last frame in which a trajectory is seen in one camera, and on the very first frame in which it is seen in the next---a rather brittle proposition.
In contrast, our measure counts the number of incorrectly matched frames, regardless of other considerations: If only one frame is wrong, the penalty is small. For instance, in the cases shown in Figure \ref{fig:handover}, current measures either charge a handover penalty when the handover is essentially correct (a) or fail to charge a handover penalty when the handover is essentially incorrect (b). Our measure charges a one-frame penalty in case (a) and a penalty nearly equal to the trajectory length in camera II in case (b), as appropriate. These cases are not just theoretical. In Section \ref{sec:experiments}, we show that 74\% of the 5,549 handovers computed by our tracker in our data set show similar phenomena.

These issues are exacerbated in measures, such as MCTA, that combine measures of within-camera mismatches and handover mismatches into a single value by a product (Eq.~\ref{eq:MCTA}). 
If one of the anomalies discussed above changes a within-camera error into a handover error or \textit{vice versa}, the corresponding contribution to the performance measure can change drastically, because the penalty moves from one term of the product to another: If the product has the form $wh$ (``within'' and ``handover''), then a unit contribution to $w$ has value $h$ in the product, and changing that contribution from $w$ to $h$ changes its value to $w$.

\subsection{The Truth-To-Result Match}
\label{sec:truth-to-result}
To address these issues, we propose to measure performance not by \emph{how often} mismatches occur, but by \emph{how long} the tracker correctly identifies targets. To this end, ground-truth identities are first matched to computed ones. More specifically, a bipartite match associates one ground-truth trajectory to exactly one computed trajectory by minimizing the number of mismatched frames over all the available data---true and computed. Standard measures such as precision, recall, and $F_1$-score are built on top of this truth-to-result match. These scores then measure the number of mismatched or unmatched detection-frames, regardless of where the discrepancies start or end or which cameras are involved.

To compute the optimal truth-to-result match, we construct a bipartite graph $G = (V_T, V_C, E)$ as follows. Vertex set $V_T$ has one ``regular'' node $\tau$ for each true trajectory and one ``false positive'' node $f^{+}_{\gamma}$ for each computed trajectory $\gamma$. Vertex set $V_C$ has one ``regular'' node $\gamma$ for each computed trajectory and one ``false negative'' node $f^{-}_{\tau}$, for each true trajectory $\tau$. Two regular nodes are connected with an edge $e\in E$ if their trajectories overlap in time. Every regular true node $\tau$ is also connected to its corresponding $f^{-}_{\tau}$, and every regular computed node $\gamma$ is also connected to its corresponding $f^{+}_{\gamma}$.

The cost on an edge $(\tau, \gamma)\in E$ tallies the number of false negative and false positive frames that would be incurred if that match were chosen. Specifically, let $\tau(t)$ be the sequence of detections for true trajectory $\tau$, one detection for each frame $t$ in the set $\mathcal{T}_{\tau}$ over which $\tau$ extends, and define $\gamma(t)$ for $t\in\mathcal{T}_{\gamma}$ similarly for computed trajectories. The two simultaneous detections $\tau(t)$ and $\gamma(t)$ are a \emph{miss} if they do not overlap in space, and we write
\begin{equation}
m(\tau, \gamma, t, \Delta) = 1\;.
\label{eq:misses}
\end{equation}
More specifically, when both $\tau$ and $\gamma$ are regular nodes, spatial overlap between two detections can be measured either in the image plane or on the reference ground plane in the world. In the first case, we declare a miss when the area of the intersection of the two detection boxes is less than $\Delta$ (with $0 < \Delta < 1$) times the area of the union of the two boxes. On the ground plane, we declare a miss when the positions of the two detections are more than $\Delta=1$ meter apart.
If there is no miss, we write $m(\tau, \gamma, t, \Delta) = 0$.
When either $\tau$ or $\gamma$ is an irregular node ($f^{-}_{\tau}$ or $f^{+}_{\gamma}$), any detections in the other trajectory are misses. When both $\tau$ and $\gamma$ are irregular, $m$ is undefined. We define costs in terms of binary misses, rather than, say, Euclidean distances, so that a miss between regular positions has the same cost as a miss between a regular position and an irregular one. Matching two irregular trajectories incurs zero cost because they are empty.

With this definition, the cost on edge $(\tau, \gamma)\in E$ is defined as follows:
\begin{equation}
\label{eq:cost_function}
c(\tau, \gamma, \Delta) = \underbrace{ \sum\limits_{t \in \mathcal{T}_{\tau}} m(\tau, \gamma, t, \Delta)}_{\text{False Negatives}} + \underbrace{ \sum\limits_{t \in \mathcal{T}_{\gamma}} m(\tau, \gamma, t, \Delta)}_{\text{False Positives}} \;.
\end{equation}

A minimum-cost solution to this bipartite matching problem determines a one-to-one matching that minimizes the cumulative false positive and false negative errors, and the overall cost is the number of mis-assigned detections for all types of errors. Every $(\tau, \gamma)$ match is a True Positive ID ($IDTP$). Every $(f^{+}_{\gamma}, \gamma)$ match is a False Positive ID ($IDFP$). Every $(\tau, f^{-}_{\tau})$ match is a False Negative ID ($IDFN$). Every $(f^{+}_{\gamma}, f^{-}_{\tau})$ match is a True Negative ID ($IDTN$).

The matches $(\tau, \gamma)$ in $IDTP$ imply a \emph{truth-to-result} match, in that they reveal which computed identity matches which ground-truth identity. In general not every trajectory is matched. The sets
\begin{equation}
MT = \{\tau \;|\; (\tau, \gamma) \in IDTP \} \;\;\;\text{ and }\;\;\;
MC = \{\gamma \;|\; (\tau, \gamma) \in IDTP \}
\end{equation}
contain the \emph{matched ground-truth trajectories} and \emph{matched computed trajectories}, respectively. The pairs in $IDTP$ can be viewed as a bijection between $MT$ and $MC$. In other words, the bipartite match implies functions $\gamma = \gamma_{m}(\tau)$ from $MT$ to $MC$ and $\tau = \tau_{m}(\gamma)$ from $MC$ to $MT$.

\subsection{Identification Precision, Identification Recall, and $F_1$ Score}

We use the $IDFN$, $IDFP$, $IDTP$ counts to compute identification precision ($IDP$), identification recall ($IDR$), and the corresponding $F_1$ score $IDF_1$. More specifically,


\begin{eqnarray}
    IDFN &=& \sum_{\tau\in AT} \sum\limits_{t \in \mathcal{T}_{\tau}} m(\tau, \gamma_{m}(\tau), t, \Delta)\\
    IDFP &=& \sum_{\gamma\in AC} \sum\limits_{t \in \mathcal{T}_{\gamma}} m(\tau_{m}(\gamma), \gamma, t, \Delta)\\
    IDTP &=& \sum_{\tau\in AT} \text{len}(\tau) - IDFN
    \;=\; \sum_{\gamma\in AC} \text{len}(\gamma) - IDFP
\end{eqnarray}
where $AT$ and $AC$ are all true and computed identities in $MT$ and $MC$.


\noindent\begin{tabularx}{\textwidth}{@{}XXX@{}}
\begin{equation}
IDP = \frac{IDTP}{IDTP + IDFP}
\end{equation}
&
\begin{equation}
IDR = \frac{IDTP}{IDTP + IDFN}
\end{equation}
\end{tabularx}

\begin{equation}
    IDF_1 = \frac{2\,IDTP}{2\,IDTP + IDFP + IDFN}
\end{equation}

Identification precision (recall) is the fraction of computed (ground truth) detections that are correctly identified. $IDF_1$ is the ratio of correctly identified detections over the average number of ground-truth and computed detections. ID precision and ID recall shed light on tracking trade-offs, while the $IDF_1$ score allows ranking all trackers on a single scale that balances identification precision and recall through their harmonic mean.

Our performance evaluation approach based on the truth-to-result match addresses all the weaknesses mentioned earlier in a simple and uniform way, and enjoys the following desirable properties: (1) \emph{Bijectivity:} A correct match (with no fragmentation or merge) between true identities and computed identities is one-to-one. (2) \emph{Optimality:} The truth-to-result matching is the most favorable to the tracker. (3) \emph{Consistency:} Errors of any type are penalized in the same currency, namely, the number of misassigned or unassigned frames. Our approach also handles overlapping and disjoint fields of view in exactly the same way---a feature absent in all previous measures.

\subsection{Additional Comparative Remarks}

\noindent\textbf{Measures of Handover Difficulty.} Handover errors in current measures are meant to account for the additional difficulty of tracking individuals across cameras, compared to tracking them within a single camera's field of view. If a system designer were interested in this aspect of performance, a similar measure could be based on the difference between the total number of errors for the multi-camera solution and the sum of the numbers of single-camera errors:
\begin{equation}
E_M - E_S \;\;\text{ where }\;\;
E_M = IDFP_M + IDFN_M
\;\;\text{ and }\;\;
E_S = IDFP_S + IDFN_S \;.
\end{equation}
The two errors can be computed by computing the truth-to-result mapping twice: Once for all the data and once for each camera separately (and then adding the single-camera errors together). The difference above is nonnegative, because the multi-camera solution must account for the additional constraint of consistency across cameras. Similarly, simple manipulation shows that ID precision, ID recall, and $IDF_1$ score are sorted the other way:
\[
IDP_S - IDP_M \geq 0 \;\;\;\text{,}\;\;\;
IDR_S - IDR_M \geq 0 \;\;\;\text{,}\;\;\;
F_{1S} - F_{1M} \geq 0
\]
and these differences measure how well the overall system can associate across cameras, given within-camera associations.

\noindent\textbf{Comparison with CLEAR MOT.} 
The first step in performance evaluation matches true and computed identities. In CLEAR MOT the event-based matching defines the best mapping sequentially at each frame. It minimizes Euclidean distances (within a threshold $\Delta$) between unmatched detections (true and computed) while matched detections from frame $t-1$ that are still within $\Delta$ in $t$ are preserved. Although the per-frame identity mapping is 1-to-1, the mapping for the entire sequence is generally many-to-many.

In our identity-based measures, we define the best mapping as the one which minimizes the total number of mismatched frames between true and computed IDs for the entire sequence. Similar to CLEAR MOT, a match at each frame is enforced by a threshold $\Delta$. In contrast, our reasoning is not frame-by-frame and results in an ID-to-ID mapping that is 1-to-1 for the entire sequence. 

The second step evaluates the goodness of the match through a scoring function. This is usually done by aggregating mistakes. MOTA aggregates FP, FN and $\Phi$ while we aggregate IDFP and IDFN counts. The notion of fragmentation is not present in our evaluation because the mapping is strictly 1-to-1. In other words our evaluation only checks whether every detection of an identity is explained or not, consistently with our definition of tracking. Also, our aggregated mistakes are binary mismatch counts instead of, say, Euclidean distances. This is because we want all errors to be penalized in the same currency. If we were to combine the binary IDFP and IDFN counts with Euclidean distances instead of IDTP, the unit of error would be ambiguous: We won't be able to tell whether the tracker under evaluation is good at explaining identities longer or following their trajectories closer.

\noindent\textbf{Comparison with Identity-Aware Tracking.} 
Performance scores similar to ours were recently introduced for this specific task~\cite{yu2016solution}. The problem is defined as computing trajectories for a known set of true identities from a database. This implies that the truth-to-result match is determined during tracking and not evaluation. Instead, our evaluation applies to the more general MTMC setting where the tracker is agnostic to the true identities.

\section{Data Set}

\begin{figure}
\centering
\includegraphics[width=0.9\columnwidth]{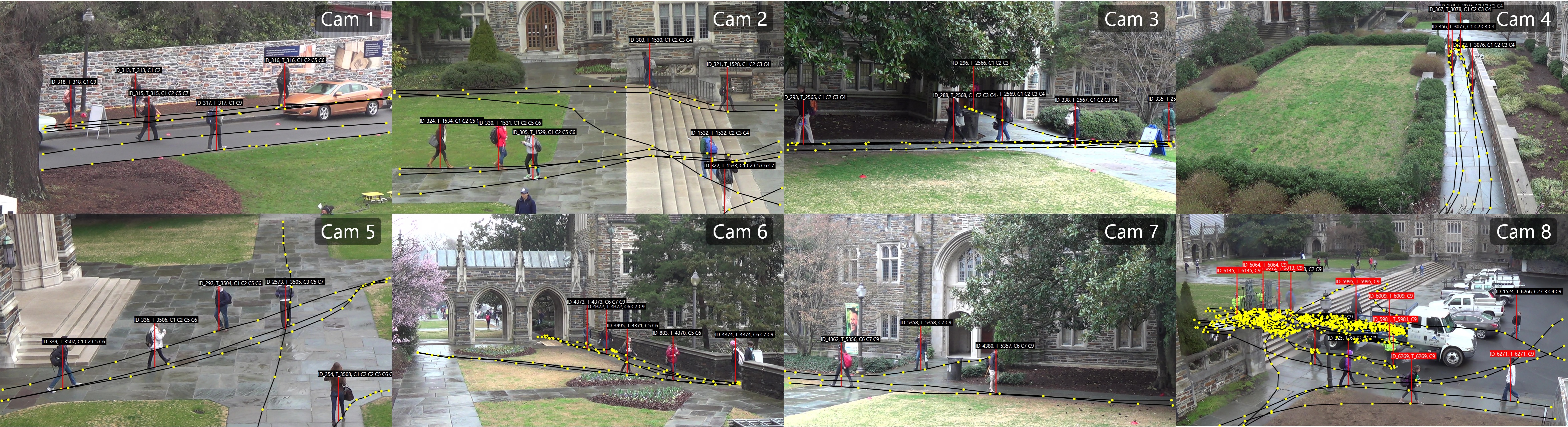}
\caption{Images and annotations of our DukeMTMC data set for frame 30890. }
\label{fig:data set}
\end{figure}

Another contribution of this work is a new, manually annotated, calibrated, multi-camera data set recorded outdoors on the Duke University campus with 8 synchronized cameras (Fig. \ref{fig:data set})\footnote{\url{http://vision.cs.duke.edu/DukeMTMC}}.
We recorded 6,791 trajectories for \emph{2,834 different identities} (distinct persons) over \emph{1 hour and 25 minutes for each camera}, for a total of more than 10 video hours and more than 2 million frames. There are on average 2.5 single-camera trajectories per identity, and up to 7 in some cases.

The cumulative trajectory time is more than 30 hours. Individual camera density varies from 0 to 54 people per frame, depending on the camera. There are 4,159 hand-overs and up to 50 people traverse blind spots at the same time. More than 1,800 self-occlusion events happen (with 50\% or more overlap), lasting 60 frames on average. Our videos are recorded at 1080p resolution and 60 fps to capture spatial and temporal detail. Two camera pairs (2-8 and 3-5) have small overlapping areas, through which about 100 people transit, while the other cameras are disjoint. Full annotations are provided in the form of trajectories of each person's foot contact point with the ground. Image bounding boxes are also available and have been semi-automatically generated. The first 5 minutes of video from all the cameras are set aside for validation or training, and the remaining 80 minutes per camera are for testing.


Unlike many multi-camera data sets, ours is not scripted and cameras have a wider field of view. Unlike single-camera benchmarks where a tracker is tested on very short videos of different challenging scenarios, our data set is recorded in a fixed environment, and the main challenge is persistent tracking under occlusions and blind spots. 


People often carry bags, backpacks, umbrellas, or bicycles. Some people stop for long periods of time in blind spots and the environment rarely constrains their paths. So transition times through blind spots are often but not always informative. 891 people walk in front of only one camera---a challenge for trackers that are prone to false-positive matches across cameras.

Working with this data set requires efficient trackers because of the amount of data to process. To illustrate, it took 6 days on a single computer to generate all the foreground masks with a standard algorithm~\cite{yao2007multi} and 7 days to generate all detections on a cluster of 192 cores using the DPM detector~\cite{5255236}. Computing appearance features for all cameras on a single machine took half a day; computing all tracklets, trajectories, and identities together also took half a day with the proposed system (Sec.~\ref{sec:system}). People detections and foreground masks are released along with the videos.

\noindent\textit{Limitations.}
Our data set covers a single outdoor scene from fixed cameras.
Soft lighting from overcast weather could make tracking easier. Views are mostly disjoint, which disadvantages methods that exploit data from
overlapping views.


\section{Reference System}
\label{sec:system}

We provide a reference MTMC tracker that extends to multiple cameras a system that was previously proposed for single camera multi-target tracking~\cite{ristani2014tracking}. Our system takes target detections from any detection system, aggregates them into tracklets that are short enough to rely on a simple motion model, then aggregates tracklets into single camera trajectories, and finally connects these into multi-camera trajectories which we call \emph{identities}.


In each of these layers, a graph $\mathcal{G} = (V, E)$ has observations (detections, tracklets, or trajectories) for nodes in $V$, and edges in $E$ connect any pairs of nodes $i, j$ for which \emph{correlations} $w_{ij}$ are provided. These are real values in $[-1, 1]$ that measure evidence for or against $i$ and $j$ having the same identity. Values of $\pm\infty$ are also allowed to represent hard evidence. A Binary Integer Program (BIP) solves the \emph{correlation clustering} problem~\cite{bansal2002correlation} on $\mathcal{G}$: Partition $V$ so as to maximize the sum of the correlations $w_{ij}$ assigned to edges that connect co-identical observations and the penalties $-w_{ij}$ assigned to edges that straddle identities. Sets of the resulting partition are taken to be the desired aggregates.

Solving this BIP is NP-hard and the problem is also hard to approximate~\cite{tan2008note}, hence the need for our multi-layered solution to keep the problems small. To account for unbounded observation times, solutions are found at all levels over a sliding temporal window, with solutions from previous overlapping windows incorporated into the proper BIP as ``extended observations''. For additional efficiency, observations in all layers are grouped heuristically into a number of subgroups with roughly consistent appearance and space-time locations.

Our implementation includes default algorithms for the computation of appearance descriptors and correlations in all layers. For appearance, we use the methods from the previous paper~\cite{ristani2014tracking} in the first layers and simple striped color histograms~\cite{liu2012person} for the last layer.  Correlations are computed from both appearance features and simple temporal reasoning.

\section{Experiments}
\label{sec:experiments}

This Section shows that (i) traditional event based measures are not good proxies for a tracker's ID precision or ID recall, defined in Section \ref{sec:performance_measures}; (ii) handover errors, as customarily defined, cause frequent problems in practice; and (iii) the performance of our reference system, when evaluated with existing measures, is comparable to that of other recent MTMC trackers. We also give detailed performance numbers for our system on our data under a variety of performance measures, including ours, to establish a baseline for future comparisons.

\begin{figure}[t]
    \centering
    \begin{tabular}{ccccccc}
        \includegraphics[width=0.23\textwidth]{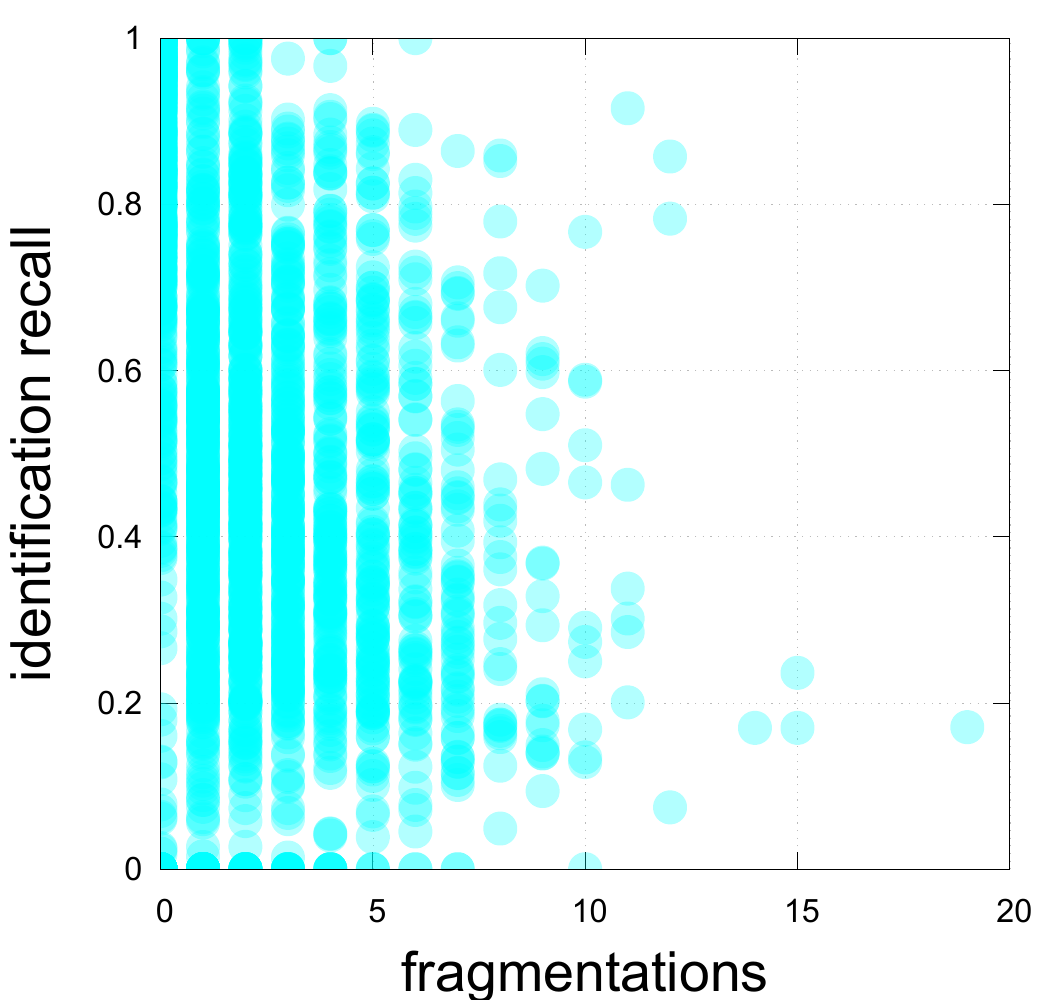} & \phantom{a} &
        \includegraphics[width=0.23\textwidth]{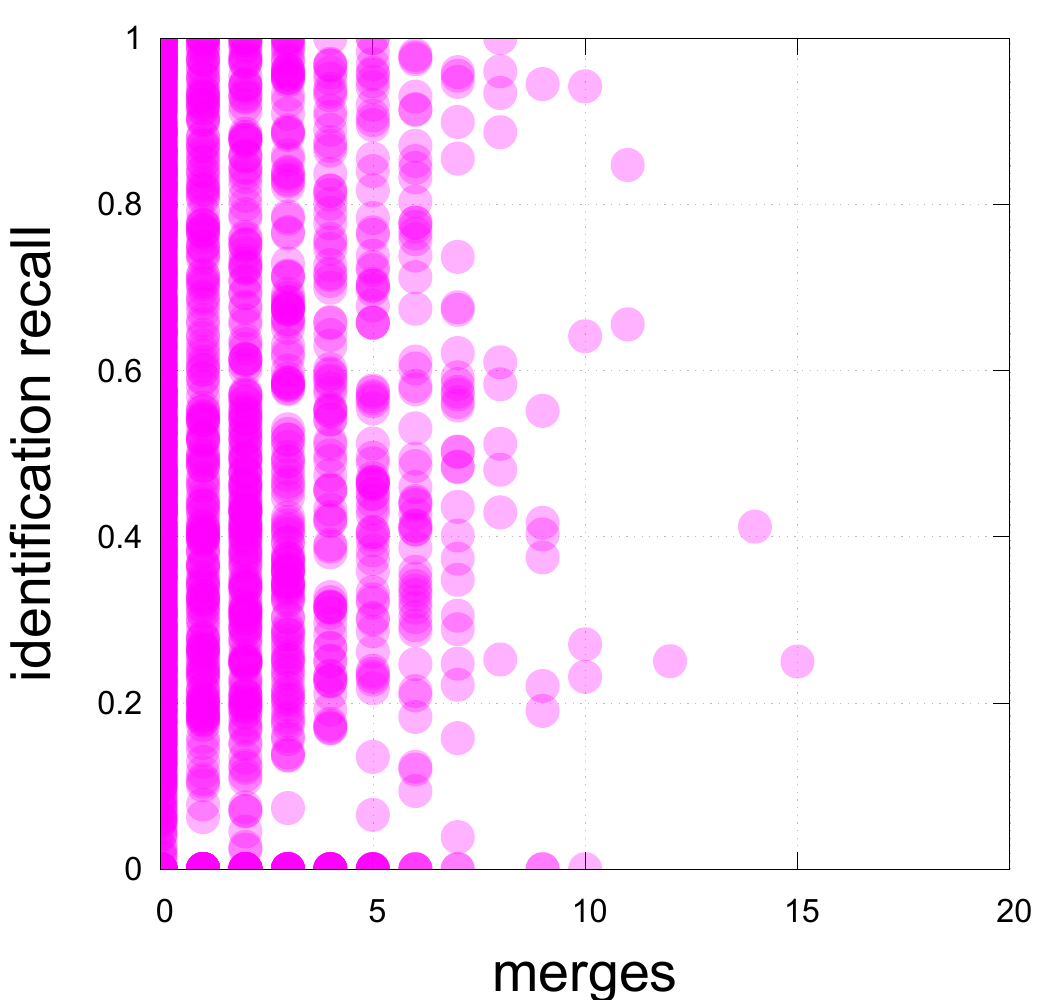} & \phantom{a} &
        \includegraphics[width=0.23\textwidth]{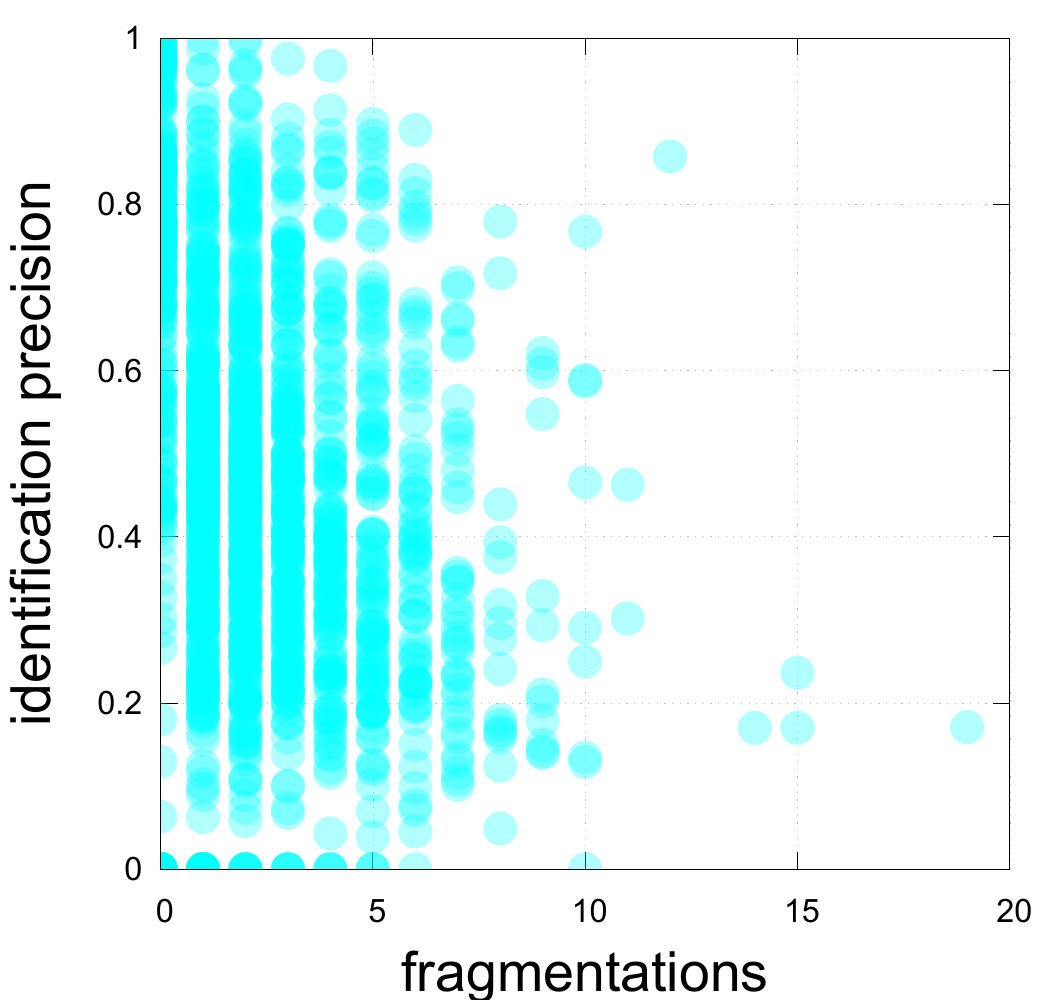} & \phantom{a} &
        \includegraphics[width=0.23\textwidth]{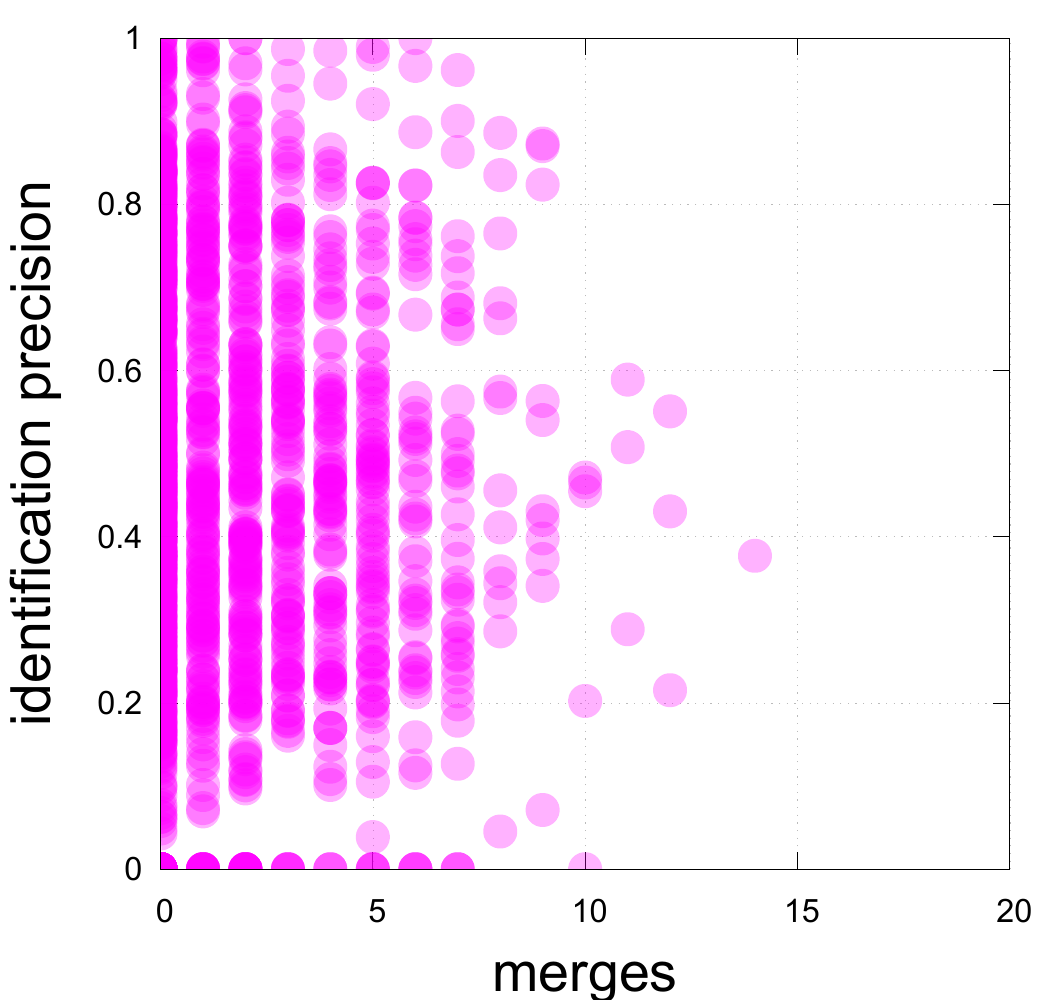} \\
        (a) & & (b) & & (c) & & (d)
    \end{tabular}
    \caption{Scatter plots of ground-truth trajectory ID recall (a, b) and ID precision (c, d) versus the number of trajectory fragmentations (a, c) and merges (b, d). Correlation coefficients are -0.24, -0.05, -0.38 and -0.41. This confirms that event- and identity-based measures quantify different aspects of tracker performance.}
    \label{fig:coverage}
\end{figure}

\noindent\textbf{ID Recall, ID Precision and Mismatches.} Figure \ref{fig:coverage} shows that fragmentations and merges correlate poorly with ID recall and ID precision, confirming that event- and identity-based measures quantify different aspects of performance.

\noindent\textbf{Truth-to-Result Mapping.} Section \ref{sec:performance_measures} and Figure \ref{fig:handover} describe situations in which traditional, event-based performance measures handle handover errors differently from ours. Figure \ref{fig:handoverCases} shows that these discrepancies are frequent in our results.

\newcommand{\enough}{\phantom{m}}
\begin{figure}[t]
    \centering
    \begin{tabular}{cccccc}
    \includegraphics[width=0.3\textwidth]{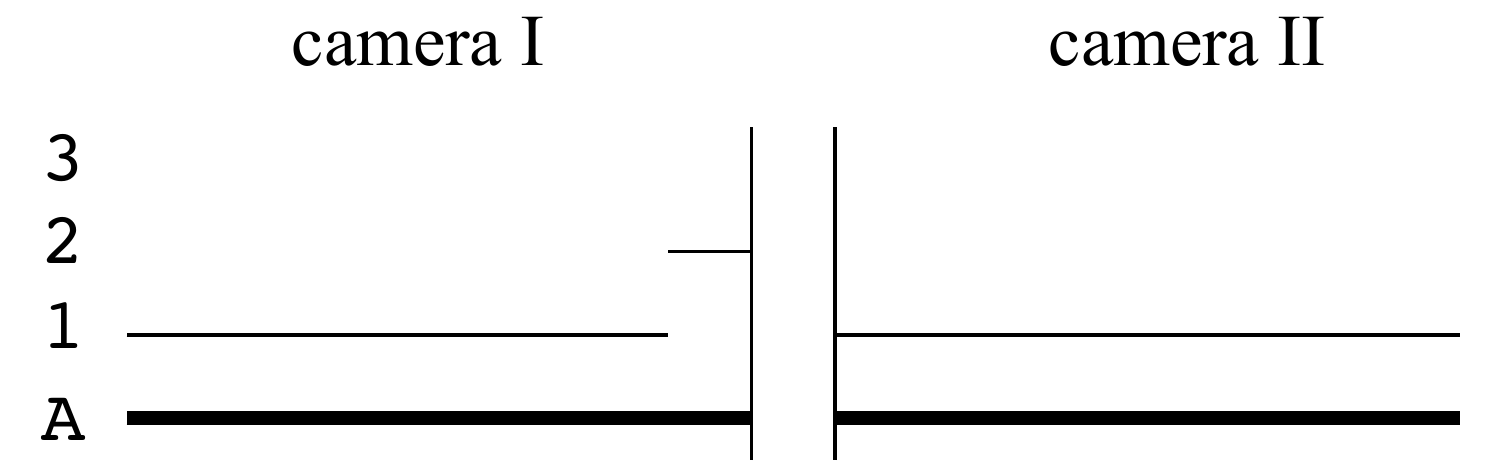} & \enough &
    \includegraphics[width=0.3\textwidth]{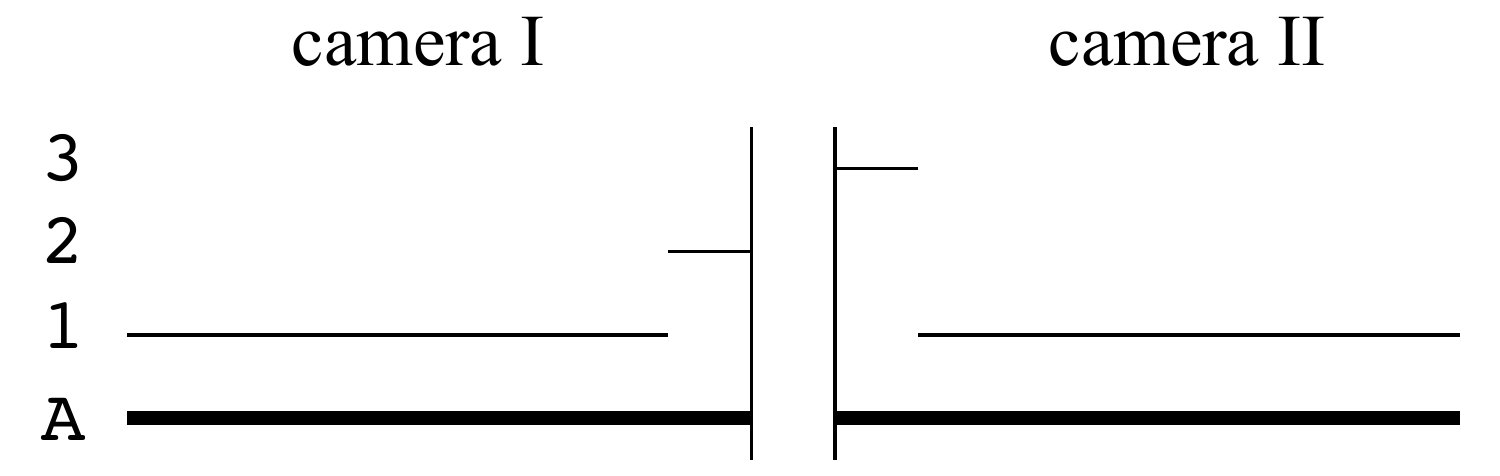} & \enough &
    \includegraphics[width=0.3\textwidth]{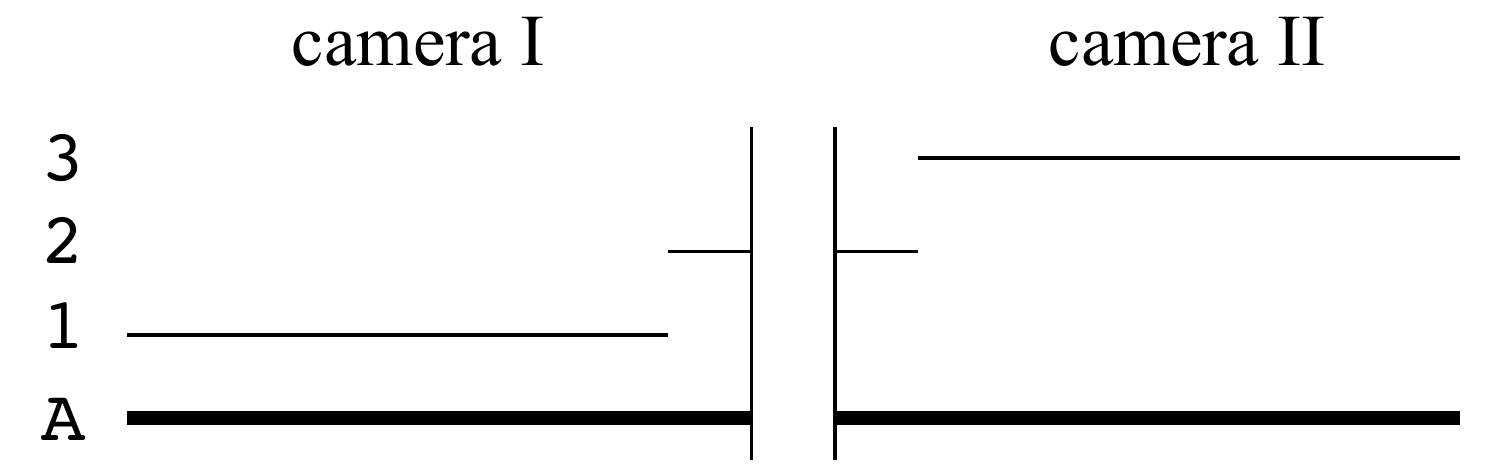} \\
    {\scriptsize (a) 1692 (30.5\%)} & &
    {\scriptsize (b) 738 (13.3\%)} & &
    {\scriptsize (c) 70 (1.3\%)} \\
    \mbox{} \\
    \includegraphics[width=0.3\textwidth]{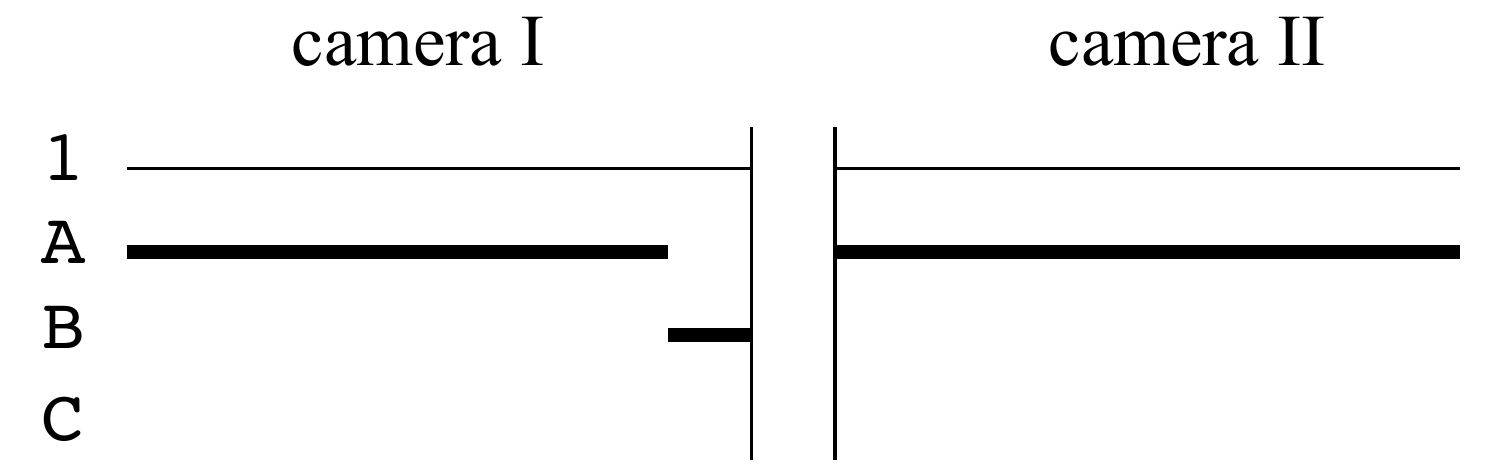} & &
    \includegraphics[width=0.3\textwidth]{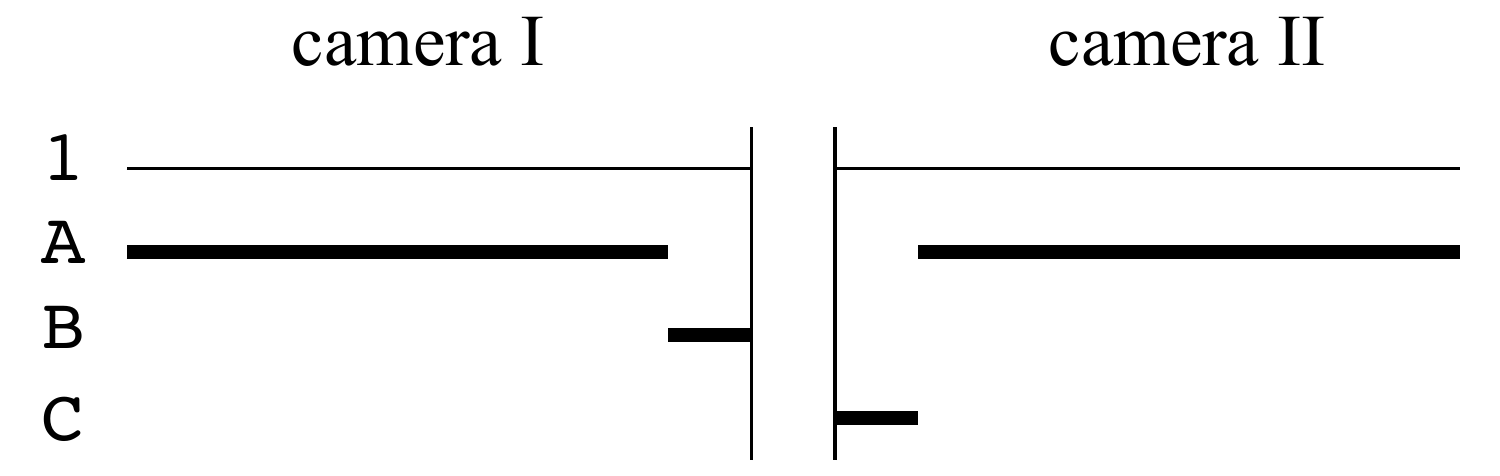} & &
    \includegraphics[width=0.3\textwidth]{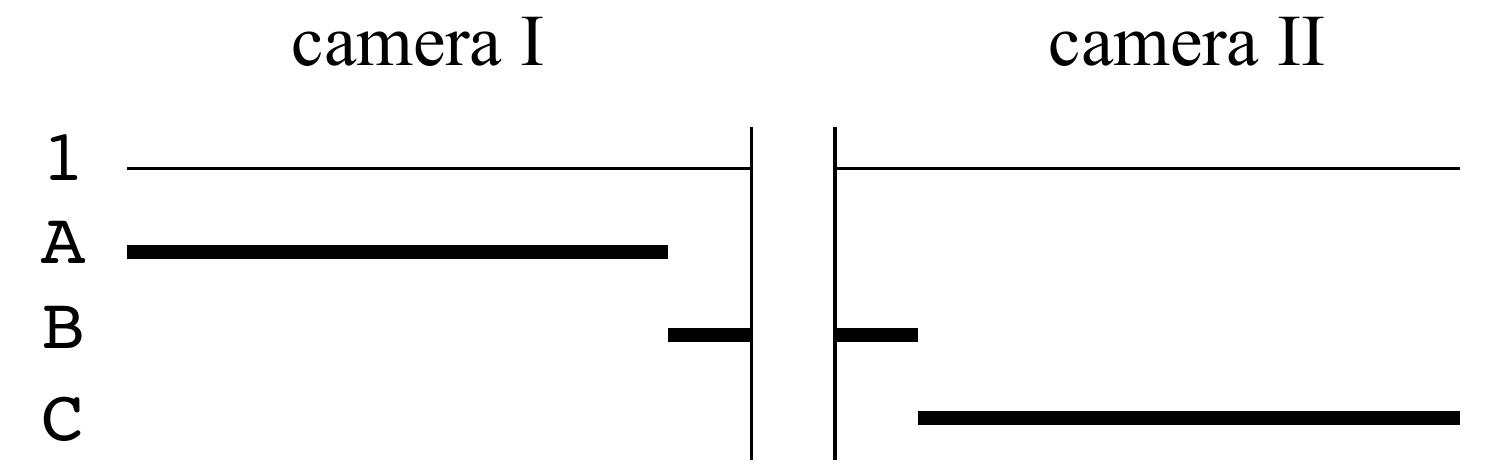} \\
    {\scriptsize (d) 1065 (19.2\%)} & &
    {\scriptsize (e) 496 (8.9\%)} & &
    {\scriptsize (f) 58 (1.0\%)}
    \end{tabular}
    \caption{[See Figure \ref{fig:handover} for the interpretation of these diagrams.] In about 74\% (4,119 out of 5,549) of the handovers output by our reference system on our data set, a short trajectory close to the handover causes a marked discrepancy between event-based, traditional performance measures and our identity-based measures. A handover fragmentation error (a, b) or merge error (d, e) is declared where the handover is essentially correct. A handover fragmentation error (c) or merge error (f) is not declared where the handover is essentially incorrect. Each caption shows the number of occurrences and the percentage of the total number of computed handovers.}
    \label{fig:handoverCases}
\end{figure}

\noindent\textbf{Traditional System Performance Analysis.} 
Table \ref{tab:final} (top) compares our reference method to existing ones on the NLPR MCT data sets~\cite{CaoCCZH15} and evaluates performance using the existing MCTA measure. The results are obtained under the commonly used experimental setup where all systems start with the same input of ground-truth single-camera trajectories. On average, our baseline system ranks second out of six by using our simple default appearance features. The highest ranked method~\cite{cai_exploring_2014} uses features based on discriminative learning.


\begin{table*}[t]
\small
\centering
\begin{tabular}{|l|c|c|c|c|c|}
\hline
\bf Systems & \bf NLPR 1 & \bf NLPR 2 & \bf NLPR 3 & \bf NLPR 4 & \bf Avg. Rank \\ \hline
USC~\cite{cai_exploring_2014} &  \textbf{0.9152} & \textbf{0.9132} &  0.5163 & 0.7052 & 2.25 \\
Ours & 0.7967 & 0.7336 & 0.6543 & \textbf{0.7616} & 2.5         \\
GE ~\cite{CaoCCZH15} & 0.8353 & 0.7034 & \textbf{0.7417} & 0.3845 & 2.75\\
hfutdspmct~\cite{mctchallenge} & 0.7425 & 0.6544 & 0.7368 & 0.3945 & 3.5 \\
CRIPAC-MCT~\cite{chen2014novel} & 0.6617 & 0.5907 & 0.7105 & 0.5703 & 4 \\
Adb-Team~\cite{mctchallenge} & 0.3204 & 0.3456 & 0.1382 & 0.1563 & 6\\
\hline 
\end{tabular}

\vspace*{3mm}
\newcommand{\sz}[1]{{\scriptsize #1}}
\begin{tabular}{|rrrrrrrrrrccc|}
\hline
                  & \multicolumn{9}{c|}{{CLEAR MOT Measures}}              &      \multicolumn{3}{c|}{{Our Measures}}                                                                                    \\
\textbf{\sz{Cam}}               & \textbf{\sz{FP$\downarrow$}} & \textbf{\sz{FN$\downarrow$}} & \textbf{\sz{IDS$\downarrow$}} & \textbf{\sz{FRG$\downarrow$}} & \textbf{\sz{MOTA$\uparrow$}} & \textbf{\sz{MOTP$\uparrow$}} & \textbf{\sz{GT}} & \textbf{\sz{MT$\uparrow$}} & \multicolumn{1}{c|}{ \textbf{\sz{ML$\downarrow$}}}  & \textbf{\sz{IDP$\uparrow$}}            & \textbf{\sz{IDR$\uparrow$}}            & \multicolumn{1}{l|}{\textbf{\sz{IDF$_1$$\uparrow$}}}            \\ \hline
\multicolumn{1}{|r|}{ 1} & 9.70 & 52.90  & 178  &  366  & 37.36  & 67.57 & 1175  & 105   & \multicolumn{1}{r|}{ 128 } & 79.17 & 44.97 & 57.36  \\ \hline 
\multicolumn{1}{|r|}{ 2} & 21.48 & 29.19 & 866  &  1929 & 49.17  & 61.70 & 1106  & 416   & \multicolumn{1}{r|}{ 50 } & 69.11 & 63.78 & 66.34  \\ \hline 
\multicolumn{1}{|r|}{ 3} & 7.04 & 39.39  & 134  &  336  & 53.50  & 63.57 & 501  & 229   & \multicolumn{1}{r|}{ 42 } & 81.46 & 55.11 & 65.74  \\ \hline 
\multicolumn{1}{|r|}{ 4} & 10.61 & 33.42 & 107  &  403  & 55.92  & 66.51 & 390  & 128   & \multicolumn{1}{r|}{ 21 } & 79.23 & 61.16 & 69.03  \\ \hline 
\multicolumn{1}{|r|}{ 5} & 3.48 & 23.38  & 162  &  292  & 73.09  & 70.52 & 644  & 396   & \multicolumn{1}{r|}{ 33 } & 84.86 & 67.97 & 75.48  \\ \hline 
\multicolumn{1}{|r|}{ 6} & 38.62 & 48.21 & 1426 &  3370 & 12.94  & 48.62 & 1043  & 207   & \multicolumn{1}{r|}{ 91 } & 48.35 & 43.71 & 45.91  \\ \hline 
\multicolumn{1}{|r|}{ 7} & 8.28 & 29.57  & 296  &  675  & 62.03  & 60.73 & 678  & 373   & \multicolumn{1}{r|}{ 53 } & 85.23 & 67.08 & 75.07  \\ \hline 
\multicolumn{1}{|r|}{ 8} & 1.29 & 61.69  & 270  &  365  & 36.98  & 69.07 & 1254  & 369   & \multicolumn{1}{r|}{ 236 } & 90.54 & 35.86 & 51.37  \\ \hline
\rowcolor[HTML]{EFEFEF} 
\multicolumn{1}{|r|}{1-8} & \multicolumn{9}{r|}{Upper bound} & 72.25 & 50.96 & 59.77  \\ \hline
\rowcolor[HTML]{EFEFEF} 
\multicolumn{1}{|r|}{1-8} & \multicolumn{9}{r|}{Baseline} & 52.35 & 36.46 & 42.98 \\ \hline
\end{tabular}
\caption{\emph{Top Table:} MCTA score comparison on the existing NLPR data sets, starting from ground truth single camera trajectories. The last column contains the average dataset ranks.
\emph{Bottom Table:} Single-camera (white background) and multi-camera (grey background) results on our DukeMTMC data set. For each separate camera we report both standard multi-target tracking measures as well as our new measures. }
\label{tab:final}
\end{table*}

\setlength{\textfloatsep}{0.1cm}

\noindent\textbf{System Performance Details.} Table~\ref{tab:final} (bottom) shows both traditional and new measures of performance, both single-camera and multi-camera, for our reference system when run on our data set. This table is meant as a baseline against which new methods may be compared.

From the table we see that our $IDF_1$ score and MOTA do not agree on how they rank the sequence difficulty of cameras 2 and 3. This is primarily because they measure different aspects of the tracker. Also, they are different in the relative value differences. For example, camera 6 appears much more difficult than 7 based on MOTA, but the difference is not as dramatic when results are inspected visually or when $IDF_1$  differences are considered. 


\section{Conclusion}

We define new measures of MTMC tracking performance that emphasize correct identities over sources of error. We introduce the largest annotated and calibrated data set to date for the comparison of MTMC trackers. We provide a reference tracker that performs comparably to the state of the art by standard measures, and we establish a baseline of performance measures, both traditional and new, for future comparisons. We hope in this way to contribute to accelerating advances in this important and exciting field.

\clearpage

\bibliographystyle{splncs}
\bibliography{refs}

\end{document}